\documentclass[10pt,twocolumn,letterpaper]{article}

\usepackage[pagenumbers]{cvpr} 

\usepackage{microtype}

\usepackage{textcomp} 
\newcommand{\mmmu}{\textsc{MMMU}}
\newcommand{\mmmuval}{\textsc{MMMU}\textsubscript{val}~}

\usepackage{booktabs,tabularx}
\usepackage{array, mathtools, xfrac}
\newcolumntype{L}[1]{>{\raggedright\arraybackslash}p{#1}}

\newcommand{\tokensinline}[2]{%
    $\lceil\sfrac{H\!\times\!W}{#1^{2}}\rceil$\,{\footnotesize\textsc{#2}}%
}

\usepackage{multirow} 
\newcolumntype{M}{>{\scriptsize}l} 
\usepackage{algorithm}
\usepackage{algorithmic}

\definecolor{cvprblue}{rgb}{0.21,0.49,0.74}
\usepackage[pagebackref,breaklinks,colorlinks,allcolors=cvprblue]{hyperref}

\title{When to Think and When to Look: Uncertainty-Guided Lookback}

\author{
Jing Bi$^{1}$ \quad
Filippos Bellos$^{2}$ \quad
Junjia Guo$^{3}$ \quad
Yayuan Li$^{2}$ \quad
Chao Huang$^{1}$ \quad
Yunlong (Yolo) Tang$^{1}$ \\
Luchuan Song$^{1}$ \quad
Susan Liang$^{1}$ \quad
Zhongfei (Mark) Zhang$^{3}$ \quad
Jason J.~Corso$^{2}$ \quad
Chenliang Xu$^{1}$ \\
[6pt]
{
$^{1}$University of Rochester \quad
$^{2}$University of Michigan \quad
$^{3}$Binghamton University
}\\[4pt]
{\tt\small \{jing.bi,yunlong.tang,chenliang.xu\}@rochester.edu, \{fbellos,yayuanli\}@umich.edu} \\
{\tt\small jjcorso@umich.edu, \{jguo22,zzhang\}@binghamton.edu, \{chuang65,sliang22,lsong11\}@cs.rochester.edu} \\
}

\begin{document}
\maketitle

\begin{abstract}
Test-time “thinking” (i.e., generating explicit intermediate reasoning chains) is known to boost performance in large language models and has recently shown strong gains for large vision–language models (LVLMs). However, despite these promising results, there is still no systematic analysis of how thinking actually affects visual reasoning. We provide the first such analysis with a large-scale, controlled comparison of thinking for LVLMs, evaluating 10 variants from the InternVL3.5 and Qwen3-VL families on MMMUval under generous token budgets and multi-pass decoding. We show that more thinking is not always better: long chains often yield \emph{long-wrong} trajectories that ignore the image and underperform the same models run in standard instruct mode. A deeper analysis reveals that certain short “lookback” phrases, which explicitly refer back to the image, are strongly enriched in successful trajectories and correlate with better visual grounding. Building on this insight, we propose \emph{uncertainty-guided lookback}, a training-free decoding strategy that combines an uncertainty signal with adaptive lookback prompts and breadth search. Our method improves overall MMMU performance, delivers the largest gains in categories where standard thinking is weak, and outperforms several strong decoding baselines, setting a new state of the art under fixed model families and token budgets. We further show that this decoding strategy generalizes, yielding consistent improvements on five additional benchmarks, including two broad multimodal suites and math-focused visual reasoning datasets.
\end{abstract}
\vspace{-6mm}    
\section{Introduction}
\begin{figure}[htbp]
  \centering
  \includegraphics[width=0.8\linewidth]{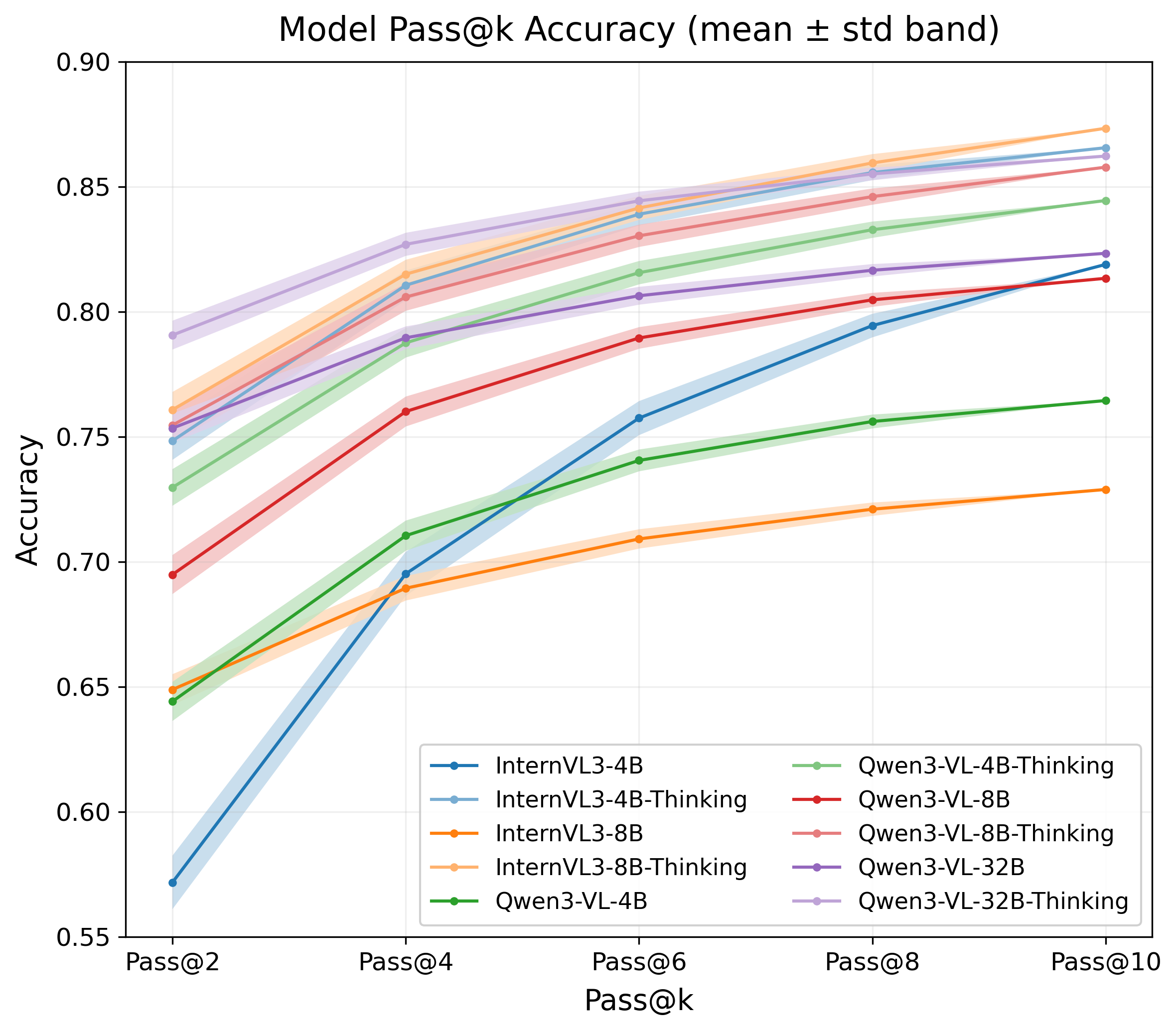}
    \caption{Pass@k accuracy on \mmmuval for 10 LVLM variants from the InternVL3.5 and Qwen3-VL families. Increasing the number of samples $k$ (breadth) yields steep early gains for all models, especially smaller ones, while enabling the built-in thinking mode consistently shifts curves upward but with diminishing returns beyond Pass@$8$. This illustrates that additional sampling can often substitute for deeper test-time thinking and that the benefits of thinking depend strongly on model capacity.}
  \label{fig:passn}
  \vspace{-6mm}
\end{figure}

LVLMs are rapidly becoming general-purpose visual assistants, expected to read charts, solve exam-style questions, and reason about diagrams at a level approaching human experts. At the same time, evaluation suites such as LMMs-Eval~\cite{zhang2024lmmseval} highlight that, despite impressive headline results, current LVLMs remain brittle across domains and task formats, motivating more principled ways of allocating test-time compute.
On the language side, \emph{Thinking}—test-time chain-of-thought decoding, self-consistency, and reflection-style prompting—has emerged as a key ingredient for complex reasoning~\cite{emmons2025whencot,tanneru2024hardnesscot}, delivering strong gains on text-only benchmarks. 
However, extending these techniques to multimodal settings is more delicate. 
Recent studies show that LVLMs inherit the failure modes of language models while adding new cross-modal issues, such as over-reliance on text priors and visual hallucination~\cite{wu2025groundedcot,zheng2024thinkingbeforelooking}, with only a few families (e.g., Qwen2.5-VL and InternVL3) exhibiting noticeably stronger visual grounding as capacity grows~\cite{zhang2025modalitypreferences}.
To better couple reasoning with perception, work on \emph{visual} CoT has begun to explicitly incorporate visual intermediates. Methods such as VCoT~\cite{rose2023visualcot}, Visual Sketchpad~\cite{hu2024visualsketchpad}, and MathCanvas~\cite{shi2025mathcanvas} endow models with drawing capabilities that act as scratchpads for geometry and spatial reasoning and thus typically rely on extra supervision or tooling.

Meanwhile, LVLM families such as Qwen-VL~\cite{bai2023qwenvl,wang2024qwen2vl,bai2025qwen25vl} and InternVL3/3.5~\cite{internvl35} expose explicit ``thinking’’ modes.
These models pair high-capacity language backbones with competitive visual encoder and report state-of-the-art results on \mmmu~ and related benchmarks. However, this trend raises three intertwined questions that are poorly understood:

\noindent \textbf{When does test-time thinking help visual reasoning?}
Thinking is widely assumed to be beneficial, but we lack systematic comparison between instruct and thinking modes across model sizes, sampling budgets, and task categories.

\noindent \textbf{How should we trade off breadth vs.\ depth of thinking?}
We do not yet know how to best allocate test-time compute between sampling more reasoning paths (breadth) and using stronger reasoning modes (depth) in perception tasks.

\noindent \textbf{Can we control thinking for better perception tasks?}
Text-only CoT work on early exit and confidence (DEER~\cite{yang2025deer}, DeepConf~\cite{fu2025deepconf}, REFRAIN~\cite{sun2025refrain}) shows that blindly elongating chains can be wasteful or harmful, motivating LVLM decoding that reacts to visual grounding and uncertainty rather than token counts alone.

This paper addresses these questions through a systematic study of visual thinking in the latest InternVL3.5~\cite{internvl35} and Qwen3-VL~\cite{bai2025qwen25vl} models, two open-source state-of-the-art LVLM families that we focus on because prior work~\cite{wu2025groundedcot,zheng2024thinkingbeforelooking} suggests they are the only ones to exhibit strong visual reasoning.
Through controlled experiments and fine-grained analysis, we show that test-time thinking has a highly structured, non-uniform impact: its benefits depend strongly on model capacity, sampling budget, and task category. In particular, we find clear regimes where thinking helps and others where concise instruct decoding is preferable.
Building on these observations, we propose \emph{uncertainty-guided lookback}, a training-free decoding strategy. Instead of always “thinking more,” we monitor when the model’s ongoing chain is likely drifting and selectively trigger short, visually anchoring lookback prompts that refocus the reasoning on the image. This mechanism, optionally combined with a lightweight parallel exploration of more visually grounded continuations, turns our analysis of long-wrong vs.\ quiet-wrong behavior into a practical recipe for routing deliberation to the instances where visual thinking actually improves.
Overall, this work makes the following contributions:

\noindent \textbf{A systematic analysis of visual thinking in LVLMs.} We provide a
    large-scale study of thinking vs.\ instruct comparisons for the latest models, disentangling breadth vs.\ depth of thinking and characterizing their effects across model sizes, categories, and difficulty levels.

\noindent \textbf{A capacity-regularized token economy for visual reasoning.} We
    show how language capacity and task difficulty jointly shape the cost and
    utility of thinking, exposing compute-equivalence trade-offs and
    highlighting when ``long-wrong'' versus ``quiet-wrong'' failures dominate.

\noindent \textbf{Uncertainty-guided lookback for adaptive visual CoT.} We
    introduce a training-free, model-agnostic decoding strategy. Across
    MMMUval and five additional benchmarks, this method consistently improves
    accuracy over standard thinking while keeping or lowering the fraction of
    thinking tokens, with especially large gains in categories where naive thinking
    previously underperformed.


\section{Related Work}


\subsection{Visual Reasoning}
Visual reasoning benchmarks increasingly minimize language priors to isolate perception–reasoning
interplay. VERIFY emphasizes explanation fidelity and reports persistent gaps on
abstract visual reasoning \cite{Bi_2025_VERIFY,tang2025captionvideofinegrainedobjectcentric,Bi_2021_ICCV}. Surveys and position papers underscore
why \emph{reasoning} (not just recognition) is central in multimodal settings
and outline evaluation pitfalls
\cite{Bi_2025_WhyReasoningSurvey,Bi_2025_DiagnosingVR}. Concurrently, reasoning-centric
LVLMs show rapid progress: \emph{LLaVA-CoT} introduces staged, autonomous visual
CoT and improves across reasoning-heavy suites \cite{Xu_2024_LLaVA_CoT}; \emph{Mulberry}
equips MLLMs with o1-like step-by-step reasoning and reflection via a collective
MCTS procedure \cite{Yao_2024_Mulberry}; \emph{InternVL~3.5} reports
strong reasoning results across MMMU/MathVista with cascade RL
\cite{internvl35}; and \emph{Qwen3-VL} releases
Instruct/Thinking variants with broad coverage and improved visual reasoning \cite{Qwen3_VL_2025}.
We investigate these families to quantify where component changes translate into visual reasoning gains.
\begin{table*}[t]
\centering
\footnotesize
\setlength{\tabcolsep}{4pt}
\renewcommand{\arraystretch}{0.85}
\begin{tabular}{l l l l l l}
    \toprule \textbf{Model}  & \textbf{Variants\,$^{\dagger}$} & \textbf{LLM backbone} & \textbf{Vision encoder} & \textbf{Connector} & \textbf{Image tokens}         \\
    \midrule InternVL3\_5-4B & instruct, COT                & Qwen3 4B              & InternViT-300M (ViT/14) & MLP projector      & \tokensinline{28}{DHR tiling} \\
    InternVL3\_5-8B          & instruct, COT                & Qwen3 8B              & InternViT-300M (ViT/14) & MLP projector      & \tokensinline{28}{DHR tiling} \\
    \midrule Qwen3-VL-4B     & instruct, thinking           & Qwen3 4B              & Qwen3VLVision (ViT/16)  & DeepStack          & \tokensinline{32}{2×2 merge}  \\
    Qwen3-VL-8B              & instruct, thinking           & Qwen3 8B              & Qwen3VLVision (ViT/16)  & DeepStack          & \tokensinline{32}{2×2 merge}  \\
    Qwen3-VL-32B             & instruct, thinking           & Qwen3 32B             & Qwen3VLVision (ViT/16)  & DeepStack          & \tokensinline{32}{2×2 merge}  \\
    \bottomrule
\end{tabular}
\caption{%
\textbf{Model comparison.} \textbf{Image tokens:} \emph{Qwen3-VL} uses an effective
stride of $32$ (ViT/16 with $2{\times}2$ token merging), yielding
$\lceil (H{\times}W)/32^{2}\rceil$ tokens for an image (e.g., $1024{\times}1024 \rightarrow
1024$ tokens). \emph{InternVL3.5} uses ViT/14 features with $4{\times}$ pixel
unshuffle and \textbf{DHR} (Dynamic High-Resolution) tiling: the image is split
into $448{\times}448$ tiles; each tile contributes $\approx 256$ tokens after compression,
and an optional global thumbnail adds $+256$. For a $1024{\times}1024$ image,
$\lceil 1024/448\rceil = 3$ per side $\Rightarrow 3{\times}3=9$ tiles, so $9{\times}
256=2304$ tokens (optionally $+256$). \textbf{DeepStack} (Qwen3-VL) denotes a multi-level
ViT feature stacking/fusion module with $2{\times}2$ token merging that reduces the
visual token grid by $4{\times}$ before the LLM. \textbf{Variants:} \emph{instruct}
vs.\ \emph{COT}/\emph{thinking}. Here, \emph{COT} means the model is \emph{prompted}
to reason and may or may not emit a thinking trace. In Qwen3-VL, \emph{thinking}
mode injects a \texttt{<think>} token in the prompt, so a thinking trace is
consistently generated. In total, we compare 8 distinct model checkpoints with 10 variants in total.
}
    \label{tab:models}
    \vspace{-1.5em}
\end{table*}

\subsection{LVLM Analysis}
Complementary work analyzes how LVLMs perceive and ground visual content: \cite{Bi_2025_CVPR} reveal visual-token–specialized attention heads whose concentration correlates with visual understanding; \cite{Kang_2025_CVPR} show that a small subset of “localization heads” suffices for competitive training-free grounding; and \cite{Huang_2025_RoPE,tang2025videolmmposttrainingdeepdive} revisit multimodal positional encodings, distilling design rules that improve grounding and video understanding. Object-level hallucination is traced to specific architectural and optimization factors with targeted mitigations \cite{Jing_2025_VOH}; an inference-time head-suppression strategy links hallucination to low image-attending heads and reduces errors with minimal latency \cite{Sarkar_2025_SPIN}; and VLM-LENS provides a toolkit for probing intermediate layers and interpretability analyses for open-source VLMs \cite{Sheta_2025_VLMLens}, informing our diagnostics separating perception from reasoning.
\section{Analysis}

\label{sec:analysis}
In this section, we first describe our experimental setup, including the
datasets, models, decoding strategies, and token budgets.
We then present our empirical findings on the effects of \emph{Thinking} and
its impact on visual reasoning.

\noindent
\textbf{Dataset and Models.} We adopt \mmmuval{} as our analysis benchmark because of its diversity and its widespread use in the LVLM literature; its 30 categories provide a rich opportunity to investigate generalization, robustness, and model strengths and failure modes~\cite{yue2024mmmu}. We focus on two leading open-source VLM families and their variants, which rank among the top-performing open-source systems on the \mmmuval{} leaderboard~\cite{llmstatsmmmuval}. As shown in Table~\ref{tab:models}, both are built upon Qwen3 LLM backbones, allowing controlled comparisons across three key dimensions: (i) vision encoder architectures (InternViT vs.\ Qwen3VLVision), (ii) connector architectures (MLP vs.\ DeepStack), and (iii) reasoning modes. This alignment makes the families well-suited for representative and systematic within-family comparison~\cite{yue2024mmmu,llmstatsmmmuval,internvl35,Qwen3_VL_2025}, 
and, to the best of our knowledge, there are currently no publicly available visual CoT controllers tailored to their ``thinking'' modes, making decoding-time control a practical way to improve performance without modifying the model.

\noindent
\textbf{Multiple-pass decoding.} We adopt a multi-sample evaluation setting with 10 sampled reasoning paths per item. This follows evidence that sampling diverse answers and marginalizing across them improves multi-step reasoning~\cite{wang2022selfconsistency} and aligns with recent work on visual reasoning and CoT prompting~\cite{zhou2025mira,mahdavi2025combigraphvis}. While prior work~\cite{zhou2025mira,mahdavi2025combigraphvis} typically uses 8 passes, we use 10 for consistency across models and a stronger assessment of answer stability. Temperature and $top_p$ follow the standard \mmmuval{} leaderboard settings, varying only random seeds~\cite{zhang2024lmmseval,evalscope_2024}, whereas common evaluation toolkits default to single-sample decoding~\cite{zhang2024lmmseval,evalscope_2024,duan2024vlmevalkit}. Further sampling details are provided in the appendix.
Here, \emph{breadth} means increasing the number of sampled trajectories (pass@$k$), while \emph{depth} means enabling the model's native reasoning mode. For Qwen3-VL, depth is controlled by the official \texttt{<think>} token; for InternVL3.5, we use only the single official reasoning prompt and do not average over alternative CoT prompts.

\begin{figure}[hbp!]
    \centering
    \includegraphics[width=0.85\linewidth]{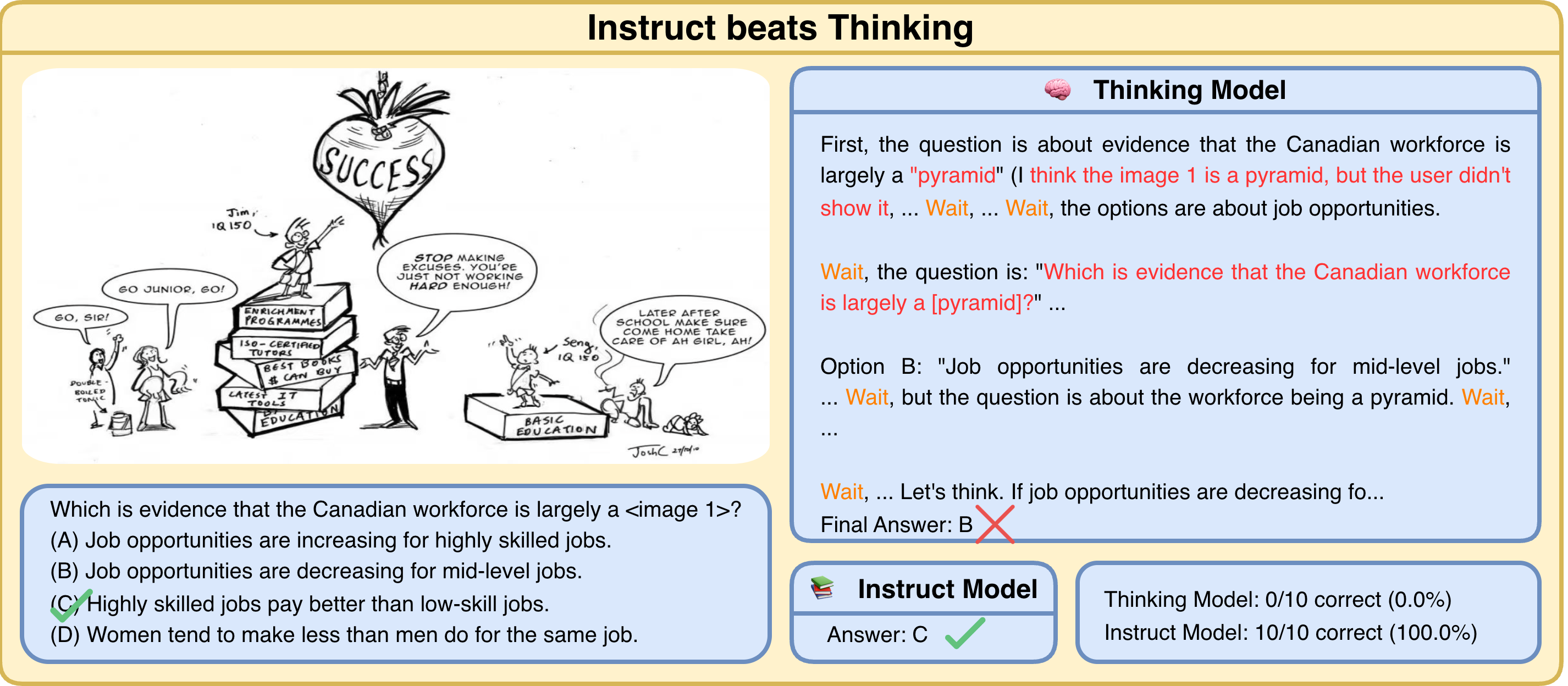}
    \caption{An example where the 32B thinking model fails on all 10 passes, while the instruct model answers correctly using 1 token.}
    \label{fig:failure}
    \vspace{-6mm}
\end{figure}
\noindent
\textbf{Token budgets.} Compared to commonly used evaluators, we employ substantially larger output-token budgets. Specifically, we allocate 16{,}384 tokens for \emph{Instruct} and 32{,}768 for \emph{Reasoning}, reducing truncation and enabling fully articulated solution chains when beneficial~\cite{zhang2024lmmseval,evalscope_2024}. This design helps ensure that performance differences are not confounded by output-length constraints (Table~\ref{tab:tokens}).
As a fairness control, we additionally run a budget-matched setting in which Thinking generations are truncated to the same 16{,}384-token cap as Instruct; the key conclusions in Secs.~\ref{sec:thinking-by-category} and \ref{sec:thinking-in-depth-breadth} remain unchanged. Full details are provided in the supplementary.
\begin{table}[t]
    \centering
    \setlength{\tabcolsep}{6pt}
    \renewcommand{\arraystretch}{0.9}
    \scriptsize

    \begin{tabularx}
        {\linewidth}{@{}>{\raggedright\arraybackslash}X c c rr@{}} \toprule
        \textbf{Evaluator} & \textbf{Year} & \textbf{Passes} & \textbf{Instruct}
        & \textbf{Reasoning} \\
        \midrule LMMs\mbox{-}Eval~\cite{zhang2024lmmseval} & 2024 & 1 & 128 & 128
        \\
        EvalScope~\cite{evalscope_2024,evalscope_docs} & 2024 & 1 & 512 & 512 \\
        VLMEvalKit~\cite{duan2024vlmevalkit} & 2024 & 1 & 512 & 512 \\
        NeMo~\cite{nemo_compute_eval} & 2025 & 1 & 2,048 & 2,048 \\
        VHELM~\cite{liang2023holistic} & 2024 & 1 & 4,096 & 4,096 \\
        Vals.ai~\cite{valsmmmuweb} & 2025 & 1 & 8,192 & 16,384 \\
        CombiGraph\mbox{-}Vis~\cite{mahdavi2025combigraphvis} & 2025 & 8 & -- &
        -- \\
        MIRA~\cite{zhou2025mira} & 2025 & 8 & 8,192 & 16,384 \\
        \midrule \textbf{Ours} & \textbf{2025} & \textbf{10} & \textbf{16,384} &
        \textbf{32,768} \\
        \bottomrule
    \end{tabularx}
    \caption{Maximum \emph{output} tokens for \mmmuval across widely used toolkits
    and recent CoT-style benchmarks, and the number of sampling passes used per
    question. We intentionally set a substantially larger budget and more passes
    to reduce truncation and enable longer reasoning chains.}
    \label{tab:tokens}
    \vspace{-2em}
\end{table}
\begin{figure*}[tbp]
    \centering
    \includegraphics[width=0.9\linewidth]{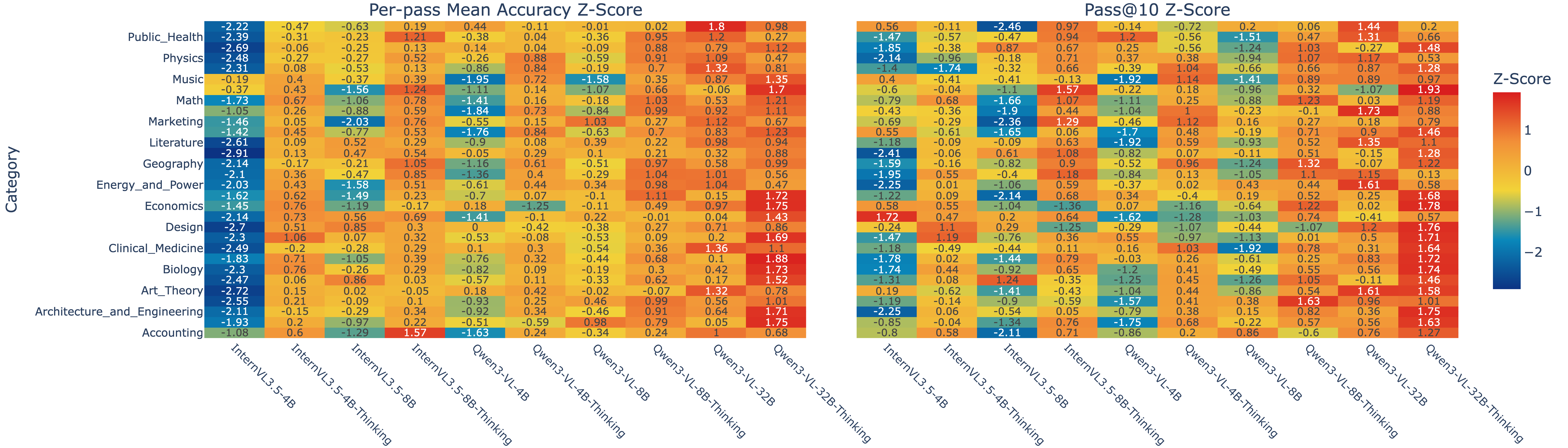}
    \caption{Category-level performance heatmaps showing z-scored accuracy across disciplines for all models and their thinking variants. Left: mean accuracy z-scores; right: Pass@10 z-scores under extensive sampling. Warmer colors indicate categories where a model performs above the global average, while cooler colors highlight relative weaknesses.}
    \label{fig:cate}
    \vspace{-6mm}
\end{figure*}
\begin{figure}[htbp]
    \centering
    \includegraphics[width=0.8\linewidth]{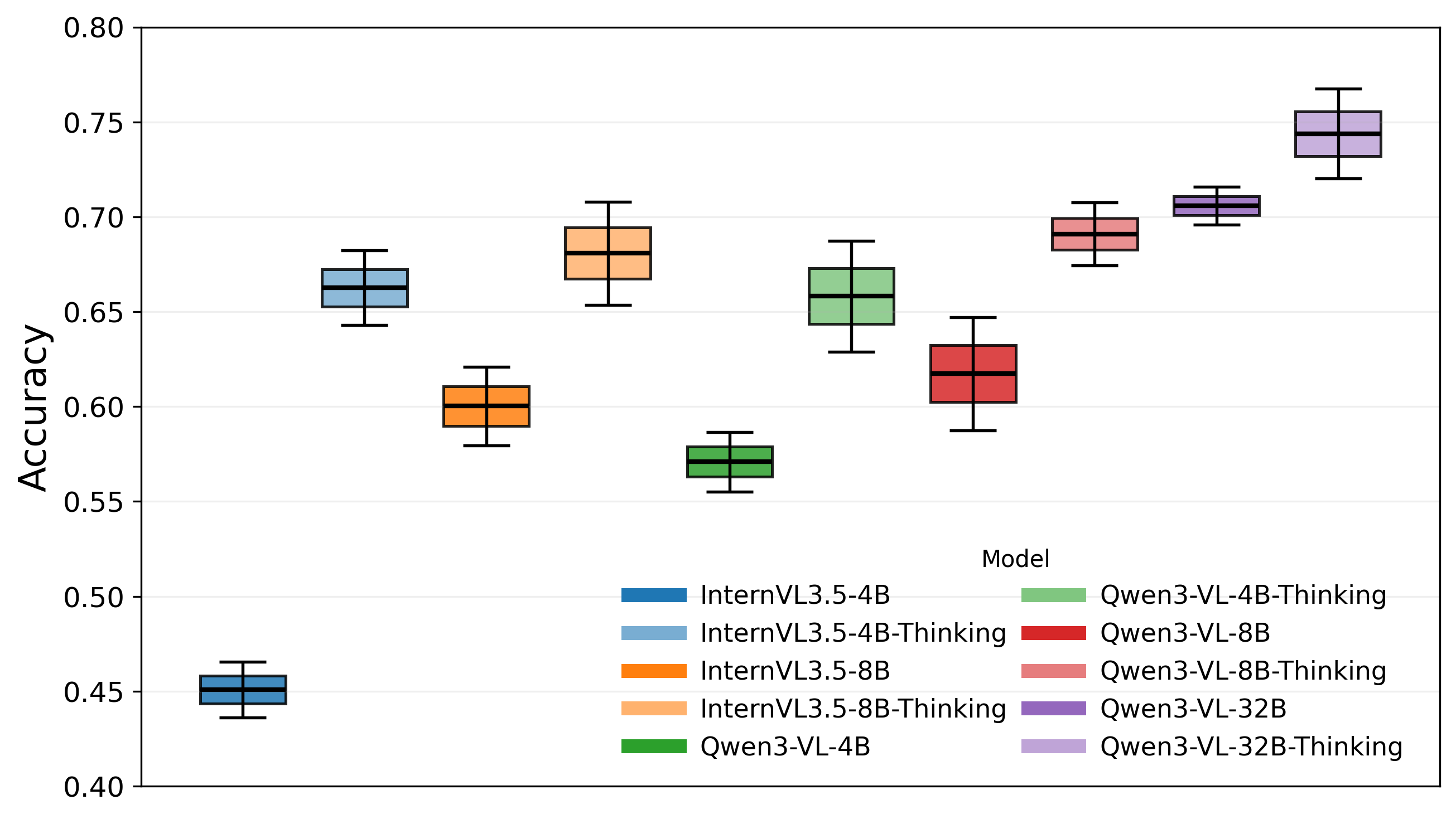}
    \caption{Mean Pass@$k$ accuracy on \mmmuval for each model and its corresponding ``Thinking'' variant. Boxplots summarize variation across evaluation runs, where higher medians and tighter interquartile ranges indicate stronger and more stable performance, highlighting differences between standard and Thinking modes.}
    \label{fig:accmean}
    \vspace{-7mm}
\end{figure}

\subsection{Will Thinking Help?}
\label{sec:thinking-in-depth-breadth}
We examine two thinking paradigms for improving reasoning: thinking in \textbf{breadth} (sampling more candidates, i.e., pass@$k$) and \textbf{depth} (explicit reasoning). The pass@$k$ curves (Fig.~\ref{fig:passn}) characterize how performance scales with additional samples, while the mean pass accuracy boxplots (Fig.~\ref{fig:accmean}) summarize the overall accuracy and variability of each model family and its thinking-enabled variant.
\vspace{-4mm}
\paragraph{Thinking in Breadth.}
Increasing $k$ monotonically improves accuracy for every model: the pass@$k$ curves in Fig.~\ref{fig:passn} steadily rise, indicating that a substantial portion of errors can be corrected with just a few extra attempts. 
The gains are steep from $k=2$ to $k=6$ and then clearly taper off for $k \ge 8$, revealing diminishing returns from pure sampling-based search.
Smaller models benefit disproportionately from larger $k$: their single-sample accuracy lags behind, but additional samples give them more chances to land on a successful reasoning trajectory.
With sufficient sampling, weaker models partially close the gap to much larger ones without explicit reasoning. For example, at high $k$ the InternVL3-4B curve moves noticeably closer to Qwen3-VL-32B (instruct), even though a performance margin remains.
In some regimes ($k \ge 6$), a smaller \emph{thinking-enabled} model can overtake a larger instruct-only model, showing that increasing $k$ can substitute for raw parameter count when latency and cost budgets allow.
\vspace{-3mm}
\paragraph{Thinking in Depth.}
Across all model sizes and values of $k$, the thinking variants consistently outperform their instruct counterparts in both Fig.~\ref{fig:passn} and Fig.~\ref{fig:accmean}. The relative gains are most pronounced for smaller models and gradually narrow as capacity increases.
Thinking improves the \emph{quality} of each sampling rather than removing the need for multiple samples: for a fixed target accuracy, thinking variants typically achieve it at a lower $k$ than their non-thinking baselines.
The standard-deviation bands in Fig.~\ref{fig:passn} and the spread of the boxplots in Fig.~\ref{fig:accmean} are generally smaller for thinking variants, suggesting more stable reasoning traces and fewer catastrophic failures across random seeds and prompts.
Thinking allows smaller models to challenge, and in some cases surpass substantially larger instruct-only models. For instance, Qwen3-VL-4B-thinking can match or exceed InternVL3.5-8B-instruct under comparable sampling budgets.
At the 8B scale, the two families largely converge once thinking is enabled: with sufficient capacity and visual modeling, additional parameter scaling yields only modest further gains, especially at higher $k$.

\noindent
\textbf{Yes, thinking helps.} Reasoning raises baseline accuracy and
reduces variance, while sampling efficiently recovers near-miss
solutions. Effective multimodal reasoning benefits from both: higher-quality
single-sample chains of thought and enough samples to explore alternatives. Nevertheless, there remain cases where even combining breadth and depth fails to resolve the problem as shown in Fig.~\ref{fig:failure}.
Additional examples are provided in the appendix.


\begin{figure*}[htbp]
    \centering
    \includegraphics[width=\linewidth]{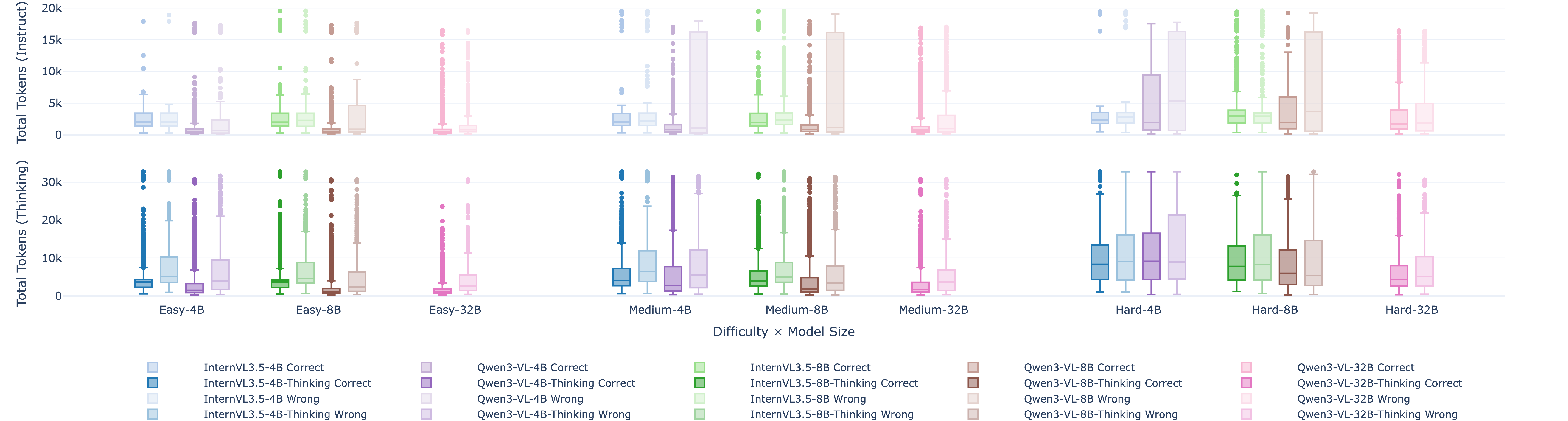}
\caption{\textbf{Token footprint of instruct vs.\ thinking across difficulty and scale.}
Boxplots show the distribution of total generated tokens per question, split by correctness (correct vs.\ wrong), model family (InternVL3.5 vs.\ Qwen3-VL), size (4B/8B/32B), and difficulty (Easy/Medium/Hard). The top panel reports \emph{Instruct} models, while the bottom reports \emph{Thinking} variants. Thinking consistently inflates token usage relative to instruct—especially on medium and hard questions and on failures—illustrating the compute overhead of deliberate reasoning. At larger scales (e.g., 32B), correct thinking traces tend to be shorter than for smaller models at the same difficulty, suggesting more efficient reasoning with increased capacity.}
    \label{fig:token}
    \vspace{-5mm}
\end{figure*}

\subsection{Does Thinking Lead in All Categories?}
\label{sec:thinking-by-category}
\noindent
Figure~\ref{fig:cate} reports per-category Z-scores for two complementary views:
\textbf{Depth} (left), the mean accuracy over 10 independent passes (single-sample
reliability), and \textbf{Breadth} (right), the aggregated accuracy under
sampling ($\mathrm{pass}@10$), i.e., the probability that at least one of $k$ samples
is correct. Within each category, scores are standardized across all model variants
(warmer~$=$~better relative to peers), and both panels share the same color
scale.

\noindent
\emph{Reasoning-centric domains favor depth for small/mid models.} In numerically
intensive categories such as Physics, Math, Engineering, Chemistry, Biology,
Clinical/Diagnostics, and computation-heavy Finance, \emph{Thinking} substantially warms the InternVL3.5-4B column in Depth relative to its \emph{Instruct} baseline, and often also improves Qwen3-VL at 8B and 32B. This mirrors the global trend from Fig.~\ref{fig:accmean}/\ref{fig:passn}: explicit reasoning makes each pass more reliable, so fewer samples are needed to reach a given accuracy. The effect is uneven across capacities and families, for instance, InternVL3.5-8B-\emph{Thinking} regresses on some STEM/business rows, suggesting that depth helps most when capacity can support coherent chains.

\noindent
\emph{Recognition-centric categories often prefer concise \emph{Instruct}.}
In Literature, History, Art/Art\_Theory, Music, and several social-science categories (Sociology, Public\_Health, Psychology), the Depth heatmap does not show a systematic warm shift for \emph{Thinking}, and in Breadth the \emph{Instruct} variants at 8B and 32B are frequently as warm as, or warmer than, their \emph{Thinking} counterparts. Here the task leans more on recognition and retrieval from priors than on multi-step deduction; longer chains mainly introduce noise, which extra sampling does not convert into more correct alternatives.


\begin{figure*}[htbp]
    \centering
    \includegraphics[width=\linewidth]{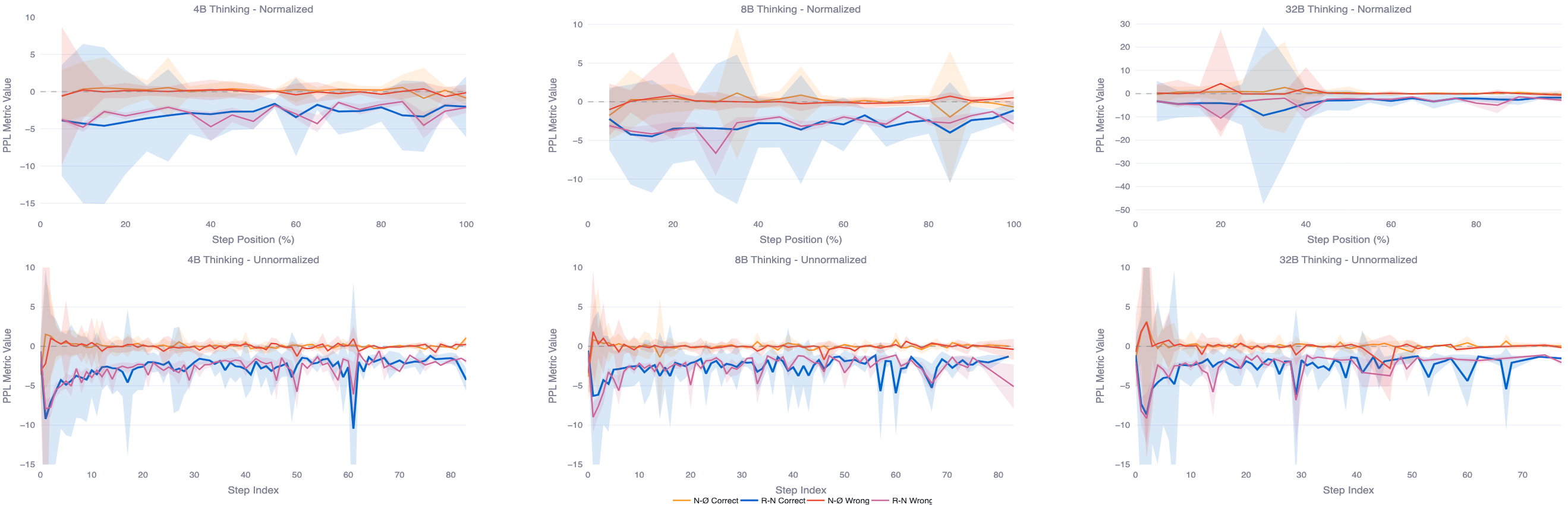}
    \caption{ Token-level $\Delta$PPL dynamics over the course of thinking for
    4B, 8B, and 32B models. Top: traces normalized to a common $0$--$100\%$ step
    position. Bottom: unnormalized step index. Blue and magenta curves
    correspond to the $R\!-\!N$ contrast for correct and wrong answers, respectively,
    while orange and red show $N\!-\!\varnothing$. The $y$-axis shows the change
    in per-token perplexity ($\Delta$PPL): negative values (e.g., $-10$) mean the
    visual condition makes the next token much easier to predict, effectively
    narrowing down the model's plausible choices, whereas positive values (e.g.,
    $+10$) mean it makes the token harder to predict, broadening or diffusing
    the set of plausible continuations. }
    \label{fig:ppl}
    \vspace{-5mm}
\end{figure*}
\subsection{How Does Thinking Perform?}
Across all difficulties (Easy/Medium/Hard), the top row of Fig.~\ref{fig:token} shows that total tokens are governed far more by \emph{language} capacity than by the vision stack or family differences. Within each difficulty group, moving from 4B$\rightarrow$8B$\rightarrow$32B systematically shortens responses and tightens the boxplots for both families, under both \emph{Instruct} and \emph{Thinking}. 

\noindent
\textbf{Capacity-driven concision is vision-stack-invariant (under a fixed LLM family).} Despite architectural
differences in the visual front-end (InternViT/14 with tiling vs.\ Qwen3-VL’s
ViT/16 tokenization), the token distributions for the two families are
remarkably similar when matched by scale. For each Easy/Medium/Hard column, the
32B variants sit lowest and most concentrated, 8B is in the middle, and 4B is
highest with the heaviest tails, in both panels. The “token economy” is
therefore largely a function of language-side capability rather than encoder or
connector design.

\noindent
\textbf{Deliberation cost is highest where it helps least.} Comparing the \emph{Instruct} and \emph{Thinking} panels, enabling reasoning roughly doubles the median token count on Easy questions at 4B and 8B, yet these are precisely the regimes where Fig.~\ref{fig:passn} shows only modest accuracy gains. On Hard questions, the multiplier is still present but less wasteful: longer chains are more likely to correspond to useful steps, especially for the smaller models. Overall, the overhead factor of \emph{Thinking} is largest for easy problems and small variants.

\noindent
\textbf{Equivalence across difficulty, capacity, and mode.} The figure implicitly traces budget frontiers: for a fixed mode, moving one step up in capacity (4B$\rightarrow$8B or 8B$\rightarrow$32B) often reduces the median token count by a similar amount to moving one step down in difficulty (Hard$\rightarrow$Medium$\rightarrow$Easy), and switching from \emph{Thinking} to \emph{Instruct} can yield a comparable saving. In practice, this enables budget-aware routing: a Hard query on a small model with \emph{Thinking} can have a similar token footprint to a Medium query on a mid-sized model or an Easy query on a large model under \emph{Instruct}, allowing systems to trade off capacity, difficulty.


\section{Improvement}
\label{sec:improvement}

In Sec.\ref{sec:analysis}, we show that \emph{more thinking is not always better}: across model size, wrong answers often use as many or more tokens than correct ones, with small models in particular producing long, low-utility chains (``long-wrong’’) on easy and medium items. These observations motivate \emph{adaptive} control over when and how to think, conditioned on both uncertainty and grounding. 
Moreover, Sec.\ref{sec:analysis} exposes systematic failures on certain \mmmuval categories, such as Sociology, where the model drifts into ungrounded textual speculation instead of using the image. Building on these diagnostics, we propose a training-free decoding strategy based on a token-level probe that (i) detects when the chain of thought enters a visually uncertain regime and (ii) triggers visual lookbacks online during streaming generation.

\subsection{Token-Level Visual Sensitivity Probe}

Following the step-decomposition protocol of~\cite{sun2025refrain}, we run each model
in thinking mode and decompose its decoded trace into word-level steps. Let the
input question be $x$, the image be $I$, and the thinking trace consists of
tokens
$
    y_{1:S}= (y_{1}, y_{2}, \dots, y_{S}).
$
For each step $s \in \{1,\dots,S\}$ we evaluate the model under three visual
contexts: \textit{Real image} $c = R$: the original image $I$; \textit{Noise image}
$c = N$: a visually mismatched image of the same resolution, e.g., Gaussian
noise that carries no semantic information about $x$ (we deliberately use synthetic noise rather than a real but unrelated image: a mismatched real image still carries rich semantic content, and LVLMs may attempt to align their reasoning with whatever objects or text it contains, confounding the probe’s interpretation of ‘no useful visual evidence’. Noise, by contrast, provides a structurally valid but semantically empty control); \textit{No image}
$c = \varnothing$: no visual tokens.

Because the decoder is autoregressive, we can compute per-step perplexity under each context to see how the image’s presence and content affect token prediction:
$
    \text{PPL}_{c}(s) = \exp\left( - \log p_{\theta}( y_{s}\mid x, I_{c}, y_{<s})
    \right),
$
where $c \in \{R, N, \varnothing\}$. We then form two difference scores:
\begin{align}
    \Delta_{\text{content}}(s)  & = \text{PPL}_{R}(s) - \text{PPL}_{N}(s), \label{eq:delta-content}            \\
    \Delta_{\text{presence}}(s) & = \text{PPL}_{N}(s) - \text{PPL}_{\varnothing}(s). \label{eq:delta-presence}
\end{align}
Intuitively, $\Delta_{\text{content}}(s)$ measures how much the \emph{correct} image content helps at step $s$. In contrast, $\Delta_{\text{presence}}(s)$ isolates the effect of merely \emph{having any image} present: a large magnitude means that the presence of visual tokens helps.
\begin{figure}[hbp!]
    \centering
    \includegraphics[width=0.95\linewidth]{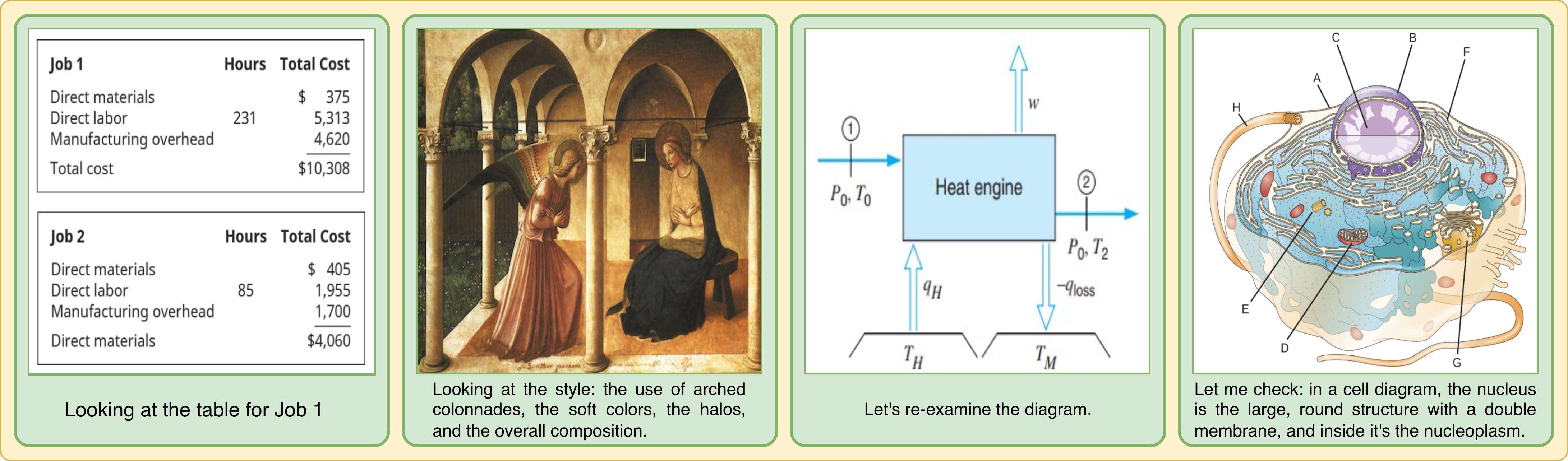}
    \caption{Examples of mined ``lookback'' sentences with strong dependence on
    the image. Each panel shows a problem image from a different domain with a reflection
    sentence that explicitly directs the model to re-examine task-relevant visual
    details.}
    \label{fig:lookback}
    \vspace{-5mm}
\end{figure}
We use the two contrast axes jointly but for different purposes. 
$\Delta_{\text{presence}}(s)$ acts as a probe for visually uncertain steps: positions with large $|\Delta_{\text{presence}}(s)|$ but small $|\Delta_{\text{content}}(s)|$ behave like generic “there is an image here’’ reactions—highly sensitive to visual tokens but not to specific content. 
By aggregating such steps across traces, we mine local $n$-grams whose usage consistently coincides with this regime and augment them with reflection-style uncertainty markers (e.g., “hmm”, “wait”) from prior work~\cite{fu2025deepconf,sun2025refrain}, yielding a compact lexical pause-phrase vocabulary that can be matched online without recomputing perplexities. 

In contrast, $\Delta_{\text{content}}(s)$ highlights strongly image-dependent reasoning: as Fig.~\ref{fig:ppl} shows, the content contrast is more negative for correctly answered examples than for wrong ones across most of the trajectory and model sizes, indicating that successful solutions maintain sustained visual grounding rather than relying on a single early glance.
In the unnormalized plots, this appears as frequent sharp dips in the correct ($R-N$) curves but fewer, shallower dips for wrong traces, indicating that successful solutions are marked by repeated, localized lookbacks to image content throughout the chain of thought.
We leverage this cue to mine \emph{lookback phrases}: multi-token templates that consistently co-occur with highly negative $\Delta_{\text{content}}(s)$ in correctly solved examples. These phrases, shown in Fig.~\ref{fig:lookback}, explicitly prompt the model to re-examine task-relevant visual details.
We mine uncertainty phrases on the same validation set used for the visual probe
(which serves as ‘pseudo ground truth’), filtering high-scoring tokens and searching
them across 10 sampled passes. We find a high alignment between mined phrases and uncertainty positions. 

\subsection{Lookback-When-Uncertain Decoding}
At test time, we implement \emph{lookback-when-uncertain} as a lightweight
controller over standard autoregressive decoding. Let $\mathcal{P}$ denote the pause-phrase
vocabulary (mined from steps with large $|\Delta_{\text{presence}}(s)|$ and
small $|\Delta_{\text{content}}(s)|$, plus uncertainty markers from~\cite{fu2025deepconf,sun2025refrain}),
and let $\mathcal{L}$ denote the set of lookback templates (extracted from steps
with highly negative $\Delta_{\text{content}}(s)$ in correct traces). During
streaming, the controller operates as:
At step $t+1$ we sample $y_{t+1}\sim p_{\theta}(\cdot \mid x, I, y_{\le t})$ and
set
\begin{equation}
    y_{1:t+1}' =
    \begin{cases}
        y_{1:t}\Vert y_{t+1}\Vert \ell, & \text{if }\substack{\neg \text{ans}(t),\, \neg \text{trig}(t),\\ \text{suffix}_{L}(y_{1:t+1}) \in \mathcal{P}}, \\[4pt]
        y_{1:t}\Vert y_{t+1},           & \text{otherwise},
    \end{cases}
\end{equation}

where $\Vert$ denotes concatenation, $\ell \in \mathcal{L}$ is a lookback phrase
(e.g., “Looking back at the image, ...”), $\text{suffix}
_{L}$ returns the length-$L$ suffix of the trace, $\text{trig}(t)$ indicates a
lookback was triggered in the last $L$ thinking tokens, and $\text{ans}(t)$ that
the model has entered the final-answer segment.

Intuitively, whenever the recent context contains a pause phrase from $\mathcal{P}$ and the model is still in the thinking phase, we immediately append a lookback template from $\mathcal{L}$, forcing an explicit re-consultation of the image before reasoning proceeds. To prevent degeneration, we allow at most one lookback trigger within any window of $L$ thinking tokens and disable further triggers once the final-answer phase starts.
All heavy computation (estimating $\Delta_{\text{presence}}(s)$ and $\Delta_{\text{content}}(s)$ and mining $\mathcal{P}$, $\mathcal{L}$) is done offline. At inference time, the controller reduces to efficient $n$-gram matching over streamed tokens plus occasional insertion of short lookback prompts, making it compatible with token-by-token streaming and avoiding any perplexity estimation.

\subsection{Parallel Lookback Sampling}

The same probe can also help choose \emph{which} visual reasoning branch to
follow. When a lookback is triggered at step $s$, we sample $M$ short
continuations of horizon $H$ after injecting $\ell$:
$y_{s:s+H}^{(m)}, \quad m = 1,\dots,M.$
For each branch we compute an aggregate visual helpfulness score
\vspace{-2mm}
\begin{equation}
    \mathcal{V}^{(m)}= -\frac{1}{H}\sum_{t = s}^{s+H-1}\Delta_{\text{content}}^{(m)}
    (t),
    \vspace{-2mm}
\end{equation}

so that larger $\mathcal{V}^{(m)}$ corresponds to trajectories where the real image
consistently reduces the loss compared to noise. We then select the branch with
maximal $\mathcal{V}^{(m)}$ and continue decoding from its state.
Because lookback events are rare and localized, this parallel lookback sampling adds only a small overhead to the total token budget, yet substantially increases the chance that at least one branch is tightly grounded in the image. This gives a compute-efficient mechanism: smaller models gain robustness by exploring multiple grounded branches, while larger models use lookback more sparingly, primarily to correct the hardest cases.
Due to page limits, we report in the appendix additional experiments using an online perplexity-based controller, and we find that its effect is very similar to our phrase-based triggers, indicating that lightweight lexical cues are sufficient and the practical gap between the probe and the deployed controller is minor.
We provide a more detailed compute and latency characterization in the appendix, including wall-clock and throughput comparisons under different configurations, showing that our method improves the trade-off with modest overhead.
\begin{table*}
    [htbp]
    \centering
    \tiny
    \setlength{\tabcolsep}{2.4pt}
    \renewcommand{\arraystretch}{1}
    \begin{tabular}{ll| cc|cc|cc|cc|cc|cc|cc}
        \toprule Size                                         & Method           & \multicolumn{2}{c}{overall} & \multicolumn{2}{c}{Finance} & \multicolumn{2}{c}{Sociology} & \multicolumn{2}{c}{Physics} & \multicolumn{2}{c}{DLM} & \multicolumn{2}{c}{EP} & \multicolumn{2}{c}{Design} \\
        \cmidrule(lr){3-4}\cmidrule(lr){5-6}\cmidrule(lr){7-8} 
                                                              &                  & Pass@1                                  & \%Tokens                    & Pass@1                        & \%Tokens                    & Pass@1                  & \%Tokens               & Pass@1                    & \%Tokens           & Pass@1            & \%Tokens           & Pass@1            & \%Tokens           & Pass@1            & \%Tokens           \\
        \midrule \multirow{6}{*}{\textbf{4B}}                 & Original         & 67.0                                    & 100.0                       & 65.3                          & 100.0                       & 60.7                    & 100.0                  & 65.7                      & 100.0              & 52.0              & 100.0              & 63.0              & 100.0              & 59.3              & 100.0              \\
                                                              & DEER             & 53.3                                    & 40.0                        & 53.3                          & 36.7                        & 46.7                    & 40.0                   & 56.7                      & 36.7               & 43.3              & 40.0               & 50.0              & 36.7               & 50.0              & 40.0               \\
                                                              & Deepconf         & 63.3                                    & 76.7                        & 66.7                          & 76.7                        & 56.7                    & 76.7                   & 66.7                      & 76.7               & 50.0              & 73.3               & 66.7              & 76.7               & 60.0              & 76.7               \\
                                                              & REFRAIN          & 63.3                                    & 73.3                        & 63.3                          & 63.3                        & 60.0                    & 76.7                   & 66.7                      & 80.0               & 53.3              & 76.7               & 66.7              & 76.7               & 63.3              & 63.3               \\
        \cmidrule(lr){2-16}                                   & Ours (lookback)  & 69.7{\tiny(+2.7)}                       & 57.2{\tiny(-42.8)}          & 68.5{\tiny(+3.2)}             & 59.4{\tiny(-40.6)}          & 62.6{\tiny(+1.9)}       & 56.3{\tiny(-43.7)}     & 67.2{\tiny(+1.5)}         & 59.1{\tiny(-40.9)} & 57.0{\tiny(+5.0)} & 57.8{\tiny(-42.2)} & 67.5{\tiny(+4.5)} & 56.9{\tiny(-43.1)} & 61.6{\tiny(+2.3)} & 59.3{\tiny(-40.7)} \\
                                                              & Ours (+sampling) & 73.0{\tiny(+6.0)}                       & 59.5{\tiny(-40.5)}          & 71.8{\tiny(+6.5)}             & 55.0{\tiny(-45.0)}          & 64.3{\tiny(+3.6)}       & 58.3{\tiny(-41.7)}     & 71.0{\tiny(+5.3)}         & 53.0{\tiny(-47.0)} & 58.3{\tiny(+6.3)} & 58.3{\tiny(-41.7)} & 72.3{\tiny(+9.3)} & 55.0{\tiny(-45.0)} & 62.0{\tiny(+2.7)} & 55.2{\tiny(-44.8)} \\
        \midrule \multirow{6}{*}{\textbf{8B}}                 & Original         & 70.3                                    & 100.0                       & 69.0                          & 100.0                       & 63.3                    & 100.0                  & 71.7                      & 100.0              & 59.7              & 100.0              & 67.3              & 100.0              & 59.0              & 100.0              \\
                                                              & DEER             & 60.0                                    & 40.0                        & 56.7                          & 40.0                        & 53.3                    & 43.3                   & 60.0                      & 40.0               & 50.0              & 43.3               & 60.0              & 40.0               & 50.0              & 40.0               \\
                                                              & Deepconf         & 66.7                                    & 80.0                        & 66.7                          & 76.7                        & 63.3                    & 80.0                   & 66.7                      & 80.0               & 56.7              & 80.0               & 70.0              & 80.0               & 56.7              & 80.0               \\
                                                              & REFRAIN          & 70.0                                    & 83.3                        & 66.7                          & 83.3                        & 60.0                    & 83.3                   & 70.0                      & 83.3               & 60.0              & 83.3               & 66.7              & 83.3               & 56.7              & 83.3               \\
        \cmidrule(lr){2-16}                                   & Ours (lookback)  & 73.0{\tiny(+2.7)}                       & 62.1{\tiny(-37.9)}          & 76.7{\tiny(+7.7)}             & 60.2{\tiny(-39.8)}          & 65.3{\tiny(+2.0)}       & 65.0{\tiny(-35.0)}     & 76.7{\tiny(+5.0)}         & 60.9{\tiny(-39.1)} & 64.3{\tiny(+4.6)} & 62.3{\tiny(-37.7)} & 75.1{\tiny(+7.8)} & 61.2{\tiny(-38.8)} & 63.7{\tiny(+4.7)} & 64.0{\tiny(-36.0)} \\
                                                              & Ours (+sampling) & 74.2{\tiny(+3.9)}                       & 63.0{\tiny(-37.0)}          & 74.1{\tiny(+5.1)}             & 58.0{\tiny(-42.0)}          & 68.8{\tiny(+5.5)}       & 63.5{\tiny(-36.5)}     & 74.1{\tiny(+2.4)}         & 59.0{\tiny(-41.0)} & 64.9{\tiny(+5.2)} & 67.3{\tiny(-32.7)} & 75.8{\tiny(+8.5)} & 63.3{\tiny(-36.7)} & 61.0{\tiny(+2.0)} & 64.8{\tiny(-35.2)} \\
        \midrule \multirow{6}{*}{\textbf{32B}}                & Original         & 75.3                                    & 100.0                       & 67.0                          & 100.0                       & 68.3                    & 100.0                  & 70.3                      & 100.0              & 65.7              & 100.0              & 71.0              & 100.0              & 77.7              & 100.0              \\
                                                              & DEER             & 66.7                                    & 43.3                        & 56.7                          & 40.0                        & 60.0                    & 43.3                   & 60.0                      & 40.0               & 56.7              & 43.3               & 60.0              & 40.0               & 66.7              & 40.0               \\
                                                              & Deepconf         & 73.3                                    & 80.0                        & 66.7                          & 80.0                        & 66.7                    & 80.0                   & 70.0                      & 80.0               & 66.7              & 80.0               & 70.0              & 80.0               & 76.7              & 80.0               \\
                                                              & REFRAIN          & 73.3                                    & 80.0                        & 66.7                          & 80.0                        & 66.7                    & 83.3                   & 70.0                      & 83.3               & 66.7              & 80.0               & 70.0              & 83.3               & 76.7              & 83.3               \\
        \cmidrule(lr){2-16}                                   & Ours (lookback)  & 81.7{\tiny(+6.4)}                       & 66.2{\tiny(-33.8)}          & 72.5{\tiny(+5.5)}             & 61.4{\tiny(-38.6)}          & 70.6{\tiny(+2.3)}       & 62.1{\tiny(-37.9)}     & 72.5{\tiny(+2.2)}         & 64.0{\tiny(-36.0)} & 71.1{\tiny(+5.4)} & 65.2{\tiny(-34.8)} & 73.3{\tiny(+2.3)} & 65.8{\tiny(-34.2)} & 80.1{\tiny(+2.4)} & 61.3{\tiny(-38.7)} \\
                                                              & Ours (+sampling) & 79.2{\tiny(+3.9)}                       & 70.3{\tiny(-29.7)}          & 73.1{\tiny(+6.1)}             & 65.3{\tiny(-34.7)}          & 71.3{\tiny(+3.0)}       & 63.0{\tiny(-37.0)}     & 76.5{\tiny(+6.2)}         & 61.0{\tiny(-39.0)} & 72.0{\tiny(+6.3)} & 66.3{\tiny(-33.7)} & 77.3{\tiny(+6.3)} & 63.5{\tiny(-36.5)} & 84.2{\tiny(+6.5)} & 61.5{\tiny(-38.5)} \\
        \bottomrule
    \end{tabular}
    \caption{\mmmuval performance (Pass@1, token usage percentage) on
    overall and selected categories for our methods on Qwen3-VL
    models. Deltas are differences w.r.t.\ the
    corresponding Original. DLM = Diagnostics and Laboratory
    Medicine, EP = Energy and Power.}
    \label{tab:mmmu_res}
    \vspace{-2mm}
\end{table*}
\begin{table}[t]
    \centering
    \scriptsize
    \setlength{\tabcolsep}{2pt}
    \renewcommand{\arraystretch}{0.9}
    \begin{tabular}{llccccc}
        \toprule Size                 & Model            & MMBench           & MMStar            & \shortstack{MathVista} & \shortstack{MathVision} & \shortstack{MathVerse} \\
        \midrule \multirow{3}{*}{4B}  & Original         & 86.7              & 73.2              & 79.5                   & 60.0                    & 75.2                   \\
        \cmidrule(lr){2-7}            & Ours (lookback)  & 89.5{\tiny(+2.8)} & 75.0{\tiny(+1.8)} & 84.3{\tiny(+4.8)}      & 64.2{\tiny(+4.2)}       & 77.2{\tiny(+2.0)}      \\
                                      & Ours (+sampling) & 88.2{\tiny(+1.5)} & 75.7{\tiny(+2.5)} & 85.0{\tiny(+5.5)}      & 65.5{\tiny(+5.5)}       & 78.7{\tiny(+3.5)}      \\
        \midrule \multirow{3}{*}{8B}  & Original         & 87.5              & 75.3              & 77.2                   & 62.7                    & 77.7                   \\
        \cmidrule(lr){2-7}            & Ours (lookback)  & 88.7{\tiny(+1.2)} & 78.5{\tiny(+3.2)} & 79.4{\tiny(+2.2)}      & 67.9{\tiny(+5.2)}       & 78.9{\tiny(+1.2)}      \\
                                      & Ours (+sampling) & 89.8{\tiny(+2.3)} & 79.6{\tiny(+4.3)} & 79.7{\tiny(+2.5)}      & 68.3{\tiny(+5.6)}       & 79.9{\tiny(+2.2)}      \\
        \midrule \multirow{3}{*}{32B} & Original         & 90.8              & 79.4              & 83.8                   & 70.2                    & 82.6                   \\
        \cmidrule(lr){2-7}            & Ours (lookback)  & 93.6{\tiny(+2.8)} & 81.2{\tiny(+1.8)} & 85.6{\tiny(+1.8)}      & 72.0{\tiny(+1.8)}       & 84.4{\tiny(+1.8)}      \\
                                      & Ours (+sampling) & 93.9{\tiny(+3.1)} & 82.5{\tiny(+3.1)} & 85.9{\tiny(+2.1)}      & 73.3{\tiny(+3.1)}       & 84.7{\tiny(+2.1)}      \\
        \bottomrule
    \end{tabular}
    \caption{Accuracy (\%) on selected benchmarks for Qwen3-VL
    Thinking models across different sizes, comparing the Original model with our variants.}
    \label{tab:others_res}
    \vspace{-6mm}
\end{table}
\subsection{Baselines}
\label{sec:baselines}
We compare against three recent \emph{training-free} adaptive reasoning methods,
all developed for text-only CoT and thus complementary to our vision-language
setting. DEER (Dynamic Early Exit in Reasoning)~\citep{yang2025deer} proposes
a confidence-based early-exit rule over reasoning tokens. DeepConf (Deep Think
with Confidence)~\citep{fu2025deepconf} uses token-level confidence to prune and
vote over multiple CoT traces. REFRAIN~\citep{sun2025refrain} introduces a
discriminator-based early-stopping policy with instance-wise threshold
adaptation. 
Together, these baselines represent the current state of training-free adaptive CoT decoding on the language side, and we use them here as strong text-only control baselines to test whether generic early-exit signals transfer to LVLMs.
\subsection{Results}

Controller necessity is confirmed by the periodic ablation: on MMMU-val (4B), periodic insertion is consistently worse than our uncertainty-guided trigger. For Qwen3-VL, periodic lookback yields 51.1/53.8/54.5/59.1/57.7 for $n=1\ldots5$, compared with 59.3 for Original Thinking and 61.6 for ours. InternVL shows the same pattern (52.2/54.9/54.0/60.4/56.9 vs.\ 57.8 Original and 59.2 ours). This supports the claim that insertion \emph{location} matters: frequent or fixed insertion can break semantic continuity, while adaptive insertion preserves useful trajectories.
For completeness, uncertainty-only branching (without lookback phrase injection) yields smaller and less stable gains than full lookback-when-uncertain, indicating that branching alone does not reliably recover visual grounding once the chain has already drifted.

To assess generality, we evaluate on \mmmuval and five established vision–language benchmarks. MMBench~\cite{liu2023mmbench} and MMStar~\cite{chen2024mmstar} probe broad multimodal capabilities, while MathVista~\cite{lu2023mathvista}, MathVision~\cite{wang2024mathvision}, and MathVerse~\cite{zhang2024mathverse} focus on visual mathematical reasoning. Detailed \mmmuval results are reported in Tab.~\ref{tab:mmmu_res}, and cross-benchmark results in Tab.~\ref{tab:others_res}. Unless noted otherwise, tabled baselines are Thinking variants of the same checkpoints.
On \mmmuval, our method consistently improves both accuracy and efficiency over the original Qwen3-VL Thinking models across all sizes (Tab.~\ref{tab:mmmu_res}). For example, on the 4B model our lookback variant raises Pass@1 from 59.3\% to 61.6\% while using only about 57\% of the original tokens, and the sampling variant pushes Pass@1 to around 62.0\% with a similar or slightly larger reduction in token usage. At 8B and 32B, we observe comparable overall gains (roughly $+2$–$+3$ absolute points) while cutting token usage by about one third. These improvements are particularly pronounced in specialist domains such as Diagnostics and Laboratory Medicine or Energy and Power, where Pass@1 can improve by up to $+5$–$+6.5$ points with 35–40\% fewer tokens. Taken together, these results indicate that our method shifts the Pareto frontier on \mmmuval: for a fixed token budget we achieve higher accuracy, and for a fixed accuracy target we require substantially fewer tokens.

The cross-benchmark results in Tab.~\ref{tab:others_res} show that these gains generalize beyond \mmmuval. Across MMBench and MMStar, both variants consistently outperform the original models at all scales, typically improving accuracy by a few absolute points even when baselines are already strong. On the math-focused benchmarks the benefits are larger: at 4B and 8B we obtain gains of roughly $+4$–$+6$ points on MathVista and MathVision and $+2$–$+3.5$ on MathVerse, with smaller but still positive gains at 32B. This pattern suggests that our approach particularly strengthens multi-step visual mathematical reasoning while also yielding robust improvements on multimodal understanding.
We also verify transfer to InternVL3.5-Think, where our method yields MMMU/MMStar gains of +1.5/+1.2 for 4B and +3.3/+3.1 for 8B, indicating that the effect is not specific to Qwen3-VL.

Comparing the two variants, the lookback strategy already offers a favorable accuracy–efficiency trade-off and is attractive when deterministic behavior is preferred. The additional sampling step further boosts performance, especially on math-heavy benchmarks and difficult \mmmuval categories, at a modest extra token cost relative to lookback alone. Overall, our approach provides a simple, plug-and-play enhancement to Qwen3-VL Thinking models that scales well with model size: it consistently yields higher accuracy—often by several absolute points—while reducing token usage by roughly 35–45\%. The appendix includes additional qualitative examples illustrating when lookback helps (e.g., correcting missed visual details) and when it can occasionally hurt.
\vspace{-5mm}
\section{Conclusion}
We find that more “vanilla thinking” is not always better for LVLMs: long chains help only in certain settings and often produce overlong, weakly grounded reasoning on easy or recognition-heavy tasks. To fix this, we propose a training-free, uncertainty-guided lookback that injects short, image-focused prompts only when chains drift, using extra compute only where visual reasoning helps. Across diverse visual benchmarks, this adaptive decoding improves accuracy over strong baselines while often using fewer tokens.
Our current implementation requires token-level log-probabilities for probe construction and trigger mining, so applicability to closed-source without log-prob access remains limited.
\newpage
\section*{Acknowledgements}
This research was funded, in part, by the U.S. Government under ARPA-H contract 1AY2AX000062, DARPA HR0011-24-9-0429, and the Center of Excellence in Data Science, an Empire State Development-designated Center of Excellence. The views and conclusions contained in this document are those of the authors and should not be interpreted as representing the official policies, either expressed or implied, of the U.S. Government.
{
    \small
    \bibliographystyle{ieeenat_fullname}
    \bibliography{main}
}


\end{document}